\documentclass[jair,twoside,11pt,theapa]{article}
\usepackage{jair, theapa, rawfonts}
\usepackage{times}
\usepackage{latexsym}
\usepackage{amsmath}
\usepackage{bbm}
\usepackage{multirow}
\usepackage{url}
\usepackage{color}
\usepackage{epsfig,url,algorithm,algorithmic,multirow}
\usepackage{amssymb}

\usepackage{booktabs}
\usepackage{graphicx}
\usepackage{varwidth}

\usepackage{array}
\newcolumntype{L}[1]{>{\raggedright\let\newline\\\arraybackslash\hspace{0pt}}m{#1}}
\newcolumntype{C}[1]{>{\centering\let\newline\\\arraybackslash\hspace{0pt}}m{#1}}
\newcolumntype{R}[1]{>{\raggedleft\let\newline\\\arraybackslash\hspace{0pt}}m{#1}}

\renewcommand{\paragraph}[1]{\noindent\textbf{#1.}}

\newcommand{\squishlist}{
 \begin{list}{}
  { \setlength{\itemsep}{0pt}
     \setlength{\parsep}{3pt}
     \setlength{\topsep}{3pt}
     \setlength{\partopsep}{0pt}
     \setlength{\leftmargin}{3em}
     \setlength{\labelwidth}{1em}
     \setlength{\labelsep}{0.5em} } }

\ShortHeadings{Knowledge-Aware Machine Reading} 
{Nakashole \& Mitchell}
\firstpageno{1}

\begin{document}

\title{Machine Reading with Background Knowledge}

\author{\name Ndapandula  Nakashole \email ndapa@cs.cmu.edu \\
       \addr Carnegie Mellon University\\
       5000 Forbes Avenue \\
		Pittsburgh, PA, 15213
       \AND
       \name Tom M Mitchell\email tom.mitchell@cs.cmu.edu \\
       \addr Carnegie Mellon University\\
       5000 Forbes Avenue \\
		Pittsburgh, PA, 15213
       }
       


\maketitle

\begin{abstract}
Intelligent  systems capable of automatically understanding natural language text are important 
for many artificial intelligence applications including mobile phone voice assistants, computer vision,  and robotics. 
Understanding language often constitutes fitting  new information into a previously acquired view of the world.
However, many machine reading systems rely  on the  text alone to infer its meaning.
In this paper, we pursue a different approach; machine reading methods that make use of   background knowledge to facilitate language understanding.
To this end, we  have developed  two methods:
The first method addresses prepositional phrase attachment ambiguity. It uses  background knowledge within a semi-supervised machine learning algorithm that learns from both labeled and unlabeled data. This approach yields state-of-the-art results on two datasets against strong baselines;
The second method extracts relationships from compound nouns.  Our knowledge-aware method for compound noun analysis accurately extracts relationships  and significantly outperforms a baseline that does not make use of background knowledge.

\end{abstract}

\section{Introduction}
\label{Introduction}

\textit{``Our feeling is that an effective characterization of knowledge can result in a real understanding system in the not too distant future."}
These were the words of Roger  Schank and Robert Abelson more than 40 years ago \cite{conf/ijcai/SchenkA75}.

A key challenge in language understanding is that most texts are prohibitively difficult to understand in isolation. Their meaning only becomes apparent when interpreted in combination with  background knowledge that reflects a previously acquired view of the world. This is true for many tasks in language understanding, ranging from co-reference resolution,  to negation detection, to prepositional phrase attachment, and even  high-level language understanding tasks such as entity linking and relation extraction. Earlier systems relied on manually specified background knowledge. \cite{conf/ijcai/SchenkA75} introduced scripts as predefined structures that describe stereotypical sequences of events for common situations.  For example, a restaurant script 
 describes  the sequence of events that happen between the time  a customer enters a restaurant and when they leave. Such a  restaurant script can then be used  to infer that if ``Alice left a restaurant after a good meal",  then with some probability,  she paid the bill and left a tip. Scripts fall within the theme of knowledge-aware machine reading. However, it is clear that with manually specified knowledge,  we cannot hope to accumulate comprehensive background knowledge. In this paper, we  study machine reading methods that leverage  automatically generated, high volume  background knowledge.

Knowledge bases are structures for characterizing and storing world knowledge. They  have been extensively studied in the past 10 years, resulting in a plethora of  large resources containing hundreds of millions of assertions about real world entities. \cite{Auer07,Bollacker2008,MitchellCHTBCMG15,suchanek2007yago}.  We aim to build  language understanding systems that make use of  the abundant  background knowledge found in knowledge bases.

We have developed two methods for sentence level machine reading  that make use of  background knowledge:

The first method addresses  a difficult case of syntactic ambiguity caused by prepositions. Prepositions  such as ``in'', ``at'', and ``for" express important details about the where, when, and why of relations and events. However, prepositions are a major source of syntactic ambiguity and still pose problems in language analysis. In particular, they cause the problem of prepositional phrase attachment ambiguity, which  arises for example,  in cases such as  ``she caught the butterfly with the spots" vs. ``she caught the butterfly with the net". In the first case,  the preposition phrase ``with the net''  modifies the verb ``caught", while in the second case, ``with the spots" modifies the noun ``butterfly". Disambiguating  these two attachments requires knowing that butterflies can have spots,  and that a net is an instrument that can be used for catching. Our  approach  uses this type of knowledge within a semi-supervised machine learning algorithm that learns from both labeled and unlabeled data. The approach produces state-of-the-art results on two datasets and  performs significantly better than the Stanford syntactic  parser, which is commonly used in  natural language processing pipelines. 
 
The second method  exploits background knowledge to extract relationships from compound nouns. Compound nouns,   consist mostly of adjectives and nouns, they do not contain verbs.   As a result, there are many lexical variations even across compound knows that express  similar semantic information between the nouns involved. Therefore, methods that rely on  co-occurrences of lexical items are bound to be limited in the task of compound noun analysis. On the other hand, relationships such as a person's job title, nationality, or stance on a political issue are often expressed using compound nouns.  For example,  ``pro-choice Democratic gubernatorial  candidate  James Florio'',  and ``White  House spokesman Marlin Fitzwater'' are compound nouns expressing useful information.  We have developed a  knowledge-aware method for compound noun analysis which accurately extracts relationships from compound nouns.\\

\paragraph{Contributions}\\
In summary, our contributions are as follows: \textit{1)~ Knowledge-Aware Machine Reading:}  We study machine reading methods
that leverage background knowledge. While the problem of machine reading has attracted a lot of attention in recent years,
there's been  little work on machine reading with background knowledge. We show compelling results on knowledge-aware machine reading  within the context of two problems: prepositional phrase attachment, and compound noun relation extraction. 
 \textit{2)~Prepositional Phrase Attachment: }  We present  a knowledge-aware method for prepositional phrase attachment.
Previous solutions to this problem largely rely on  corpus statistics. Our approach  draws upon 
 diverse sources of background knowledge, leading to significant performance improvements. 
In addition to training on labeled  data, we also make use of a large amount of unlabeled data. This enhances our method's ability to generalize to diverse data sets. 
In addition to the standard Wall Street Journal corpus (WSJ)~\cite{Ratnaparkhi1994}, we labeled two new datasets for testing purposes, one from Wikipedia (WKP), and the other from the  New York Times Corpus (NYTC). We make  these datasets freely available for future research.  In addition, we have applied our  model to over 4 million 5-tuples of the form \textit{noun0, verb, noun1, preposition, noun2}, and we also make this dataset  available\footnote{http://rtw.ml.cmu.edu/resources/ppa}.
This work was first published in \cite{conf/acl/NakasholeM15},  in this paper we report  additional experiments on ternary relations.
We also place this work in the larger context of knowledge-aware machine reading.
\textit{3)~Compound Noun Analysis: }  We introduce a knowledge-aware method for extracting relations from compound nouns. We collected over 2 million compound nouns from which we learned fine-grained semantic type sequences that express ontological  from the NELL knowledge base. Our experiments show that we obtain significantly higher accuracy than a baseline.\\

\paragraph{Organization}\\
The rest of the paper is organized as follows.   Section \ref{ppa} presents our  method for
prepositional phrase attachment disambiguation. In addition to our main results, we also present  findings on  how we can use our method in high-level machine reading tasks such as relation extraction, in particular, ternary relation extraction.  Section \ref{nominals} introduces our  approach to  compound noun analysis.  In addition to the task-specific related work presented in each of the first two sections, Section \ref{relatedworkall} presents additional work related to  knowledge-aware machine reading. Lastly, in Section \ref{discussion} we discuss the implications of our results, and bring forward a number of open questions.

\section{Prepositional Phrase Attachment}\label{ppa}
Prepositional phrases (PPs)   express crucial information that  information extraction  methods need to extract.
 However, PPs are a major source of  syntactic  ambiguity. In this paper, we introduce an algorithm that uses background knowledge to  improve PP attachment accuracy. 
 Prepositions such as  ``in", ``at", and ``for" express details about the  \textit{where, when,} and \textit{why}  of  relations and events. PPs   also state  attributes of nouns. 

\begin{figure}[t]

\centering

\includegraphics[width=1\columnwidth] {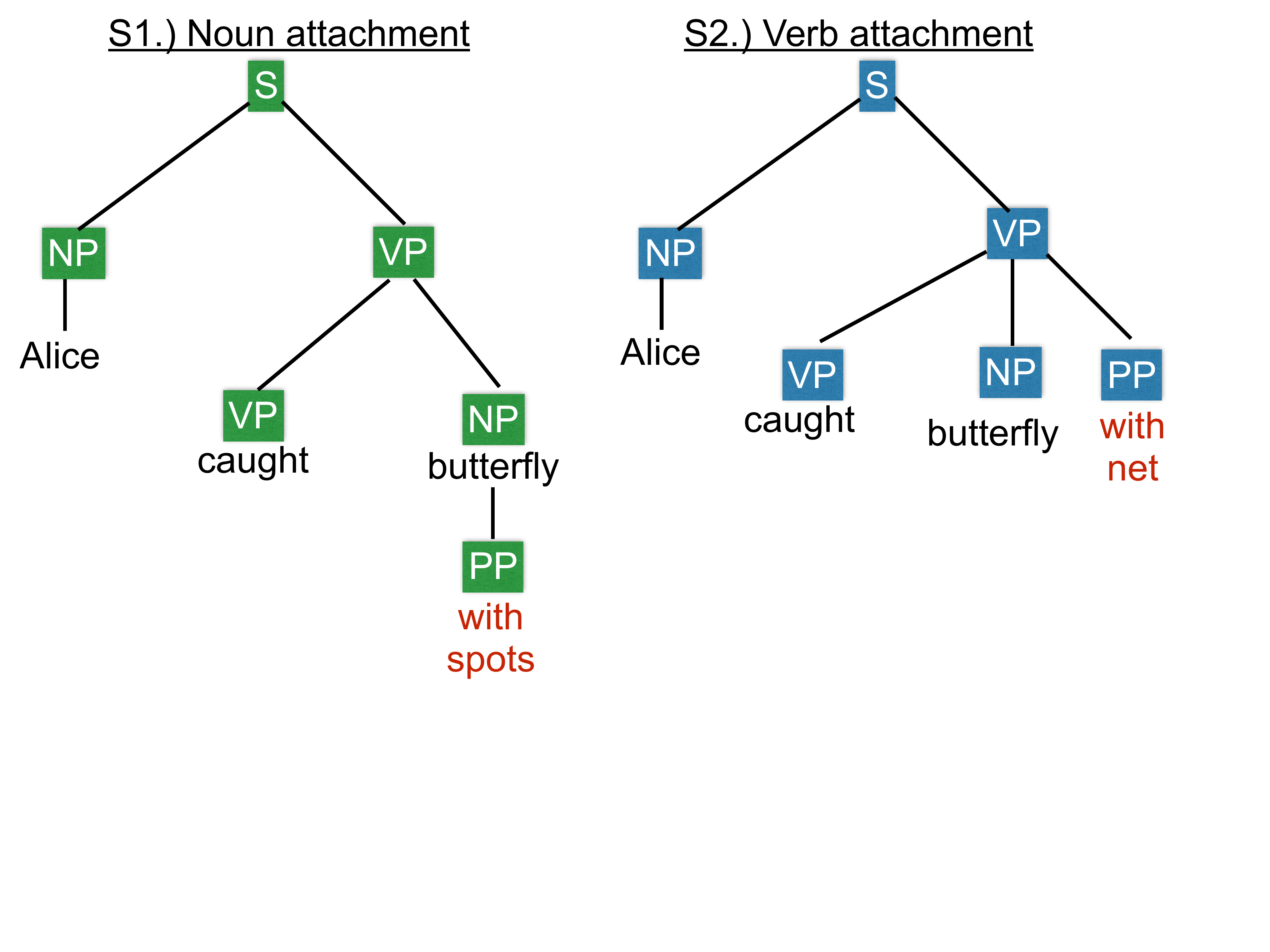}
\vspace*{-2.6cm}
\caption{Parse trees where the prepositional phrase (PP) attaches to the noun, and to the verb.}

\label{fig:deptrees}

\end{figure}

 As an example, consider the following sentences: \textit{ S1.) Alice caught the butterfly with the spots. S2.) Alice caught the butterfly with the net. }  S1 and S2  are identical except in their final nouns. However their parses differ as seen in Figure \ref{fig:deptrees}). This is because  in  S1,  the butterfly has spots and therefore  the PP, ``with the spots'', attaches to the \textit{noun}. In the task relation extraction, we  obtain a \textit{binary} relation of the form:  
 $\langle$Alice$\rangle$  caught $\langle$butterfly with  spots$\rangle$.
However, in S2, the net is the instrument used for catching and therefore  the PP,  ``with the net", attaches to the \textit{verb}.  For relation extraction, we get a \textit{ternary} extraction of the form:
 $\langle$Alice$\rangle$  caught $\langle$butterfly$\rangle$ with $\langle$net$\rangle$.
 
The PP attachment problem is often defined as follows: given a PP occurring within a  sentence where there are multiple possible attachment sites for the PP, choose the most plausible attachment site. 
 In the literature,  prior work going as far back as \cite{BrillR94,Ratnaparkhi1994,Collins95} has  focused on the  language pattern that causes most PP ambiguities, which is the  4-word sequence: $\{v, n1, p, n2\}$ (e.g., $\{${\em caught, butterfly, with, spots}$\}$). The task is  to   determine if  the prepositional phrase $(p,n2)$  attaches to  the verb $v$ or to the first noun $n1$.
Following common practice,  we focus on  PPs occurring as $\{v,n1,p,n2\}$ quadruples ---  we shall refer to these as  \textit{PP quads}. 

The approach we present here differs from prior work in two main ways. First, we make extensive use of semantic knowledge about nouns, verbs, prepositions, pairs of nouns, and  the discourse context in which a PP quad occurs. Table \ref{tab:knowledge}  summarizes the types of  knowledge we considered in our work. Second, in training our model, we rely on both labeled and unlabeled data, employing an expectation maximization (EM) algorithm \cite{Dempster77maximumlikelihood}.

\begin{table}[h]
\centering
\small{
   \begin{tabular}{|p{1.8cm}|p{4.9cm}|}
     \hline
     Relations &  Noun-Noun binary relations  \newline \textit{ (Paris, located in, France)} \newline \textit{(net, caught, butterfly)}\\
     \hline
     Nouns &  Noun semantic categories \newline \textit{(butterfly, isA, animal)}  \\
     \hline
     Verbs & Verb roles \newline  \textit{caught(agent, patient, instrument)} \\
     \hline
     Prepositions& Preposition  definitions \newline  
     \textit{ f(for)= used for, has purpose, ...}  
     \newline \textit{f(with)= has, contains, ...}  \\
     \hline
     Discourse &  Context \newline  $n0 \in \{n0, v, n1, p, n2\}$\\
     \hline
   \end{tabular}
   \caption{Types of background   
   knowledge used in this paper to determine PP attachment.}
     \label{tab:knowledge}
     }      
   \end{table}

\begin{figure}[t]
\centering
\includegraphics[width=0.80\columnwidth] {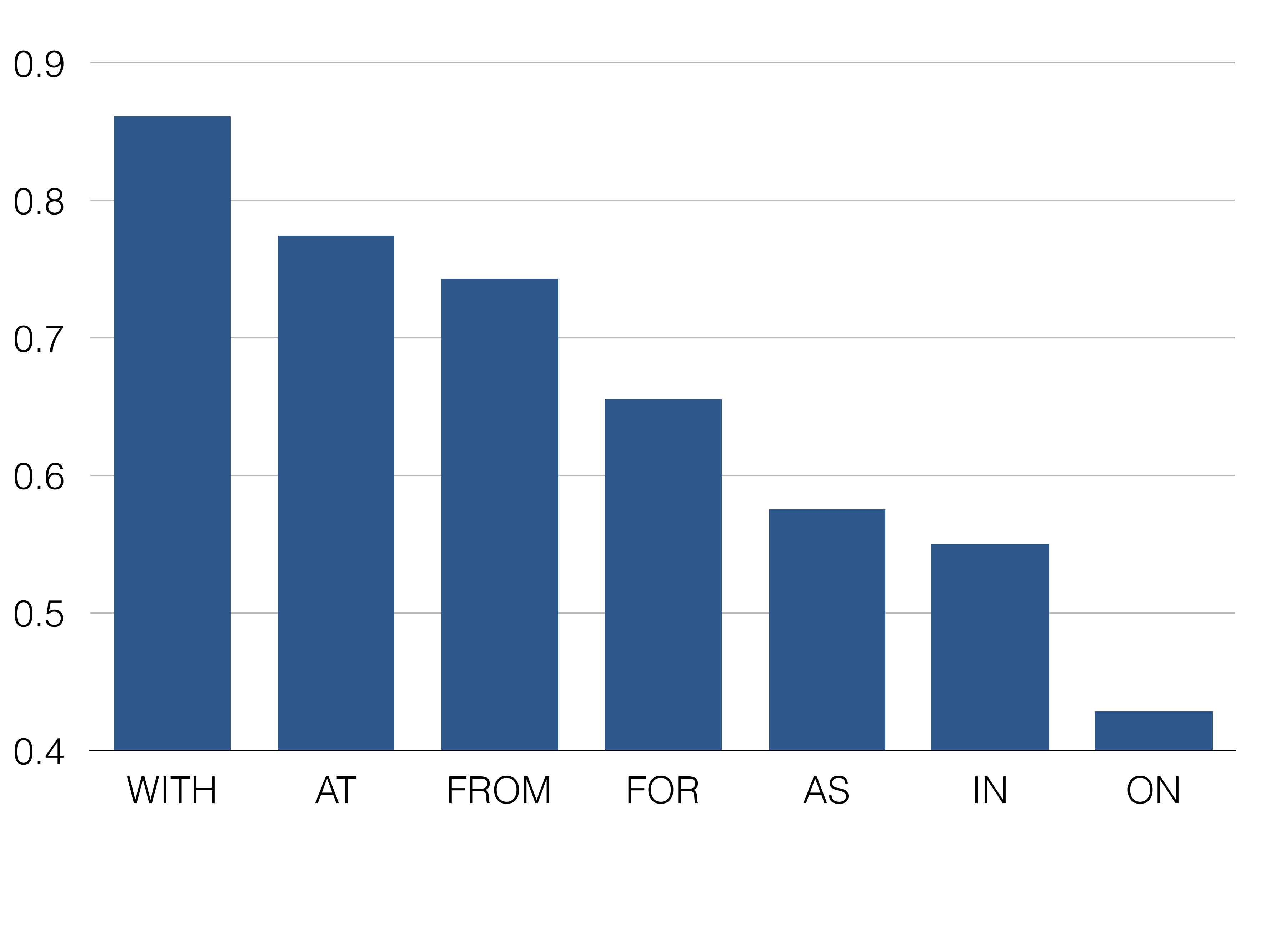}
\vspace*{-0.8cm}
\caption{Dependency parser PP attachment accuracy for various frequent prepositions.}
%
\label{fig:parser}
\end{figure}

\subsection{State  of  the Art}

\begin{figure}[t]
 \centering
 \includegraphics[width=0.85\columnwidth] {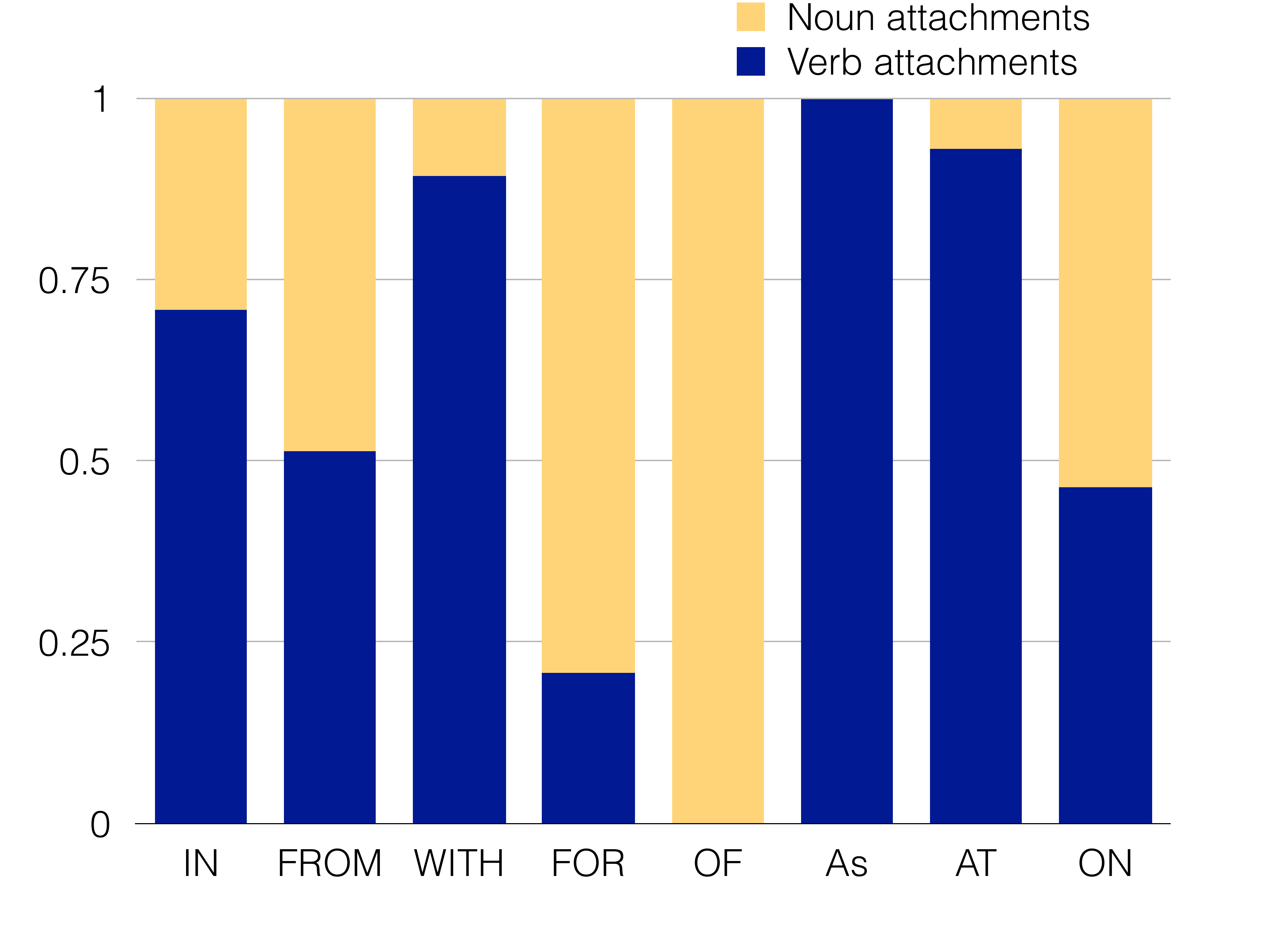}
 \vspace*{-0.5cm}
 \caption{Noun vs. verb attachment proportions for frequent prepositions in the labeled NYTC dataset.}
 %
 \label{fig:distribution}
 \end{figure}  

 To quantitatively assess existing tools, we analyzed performance of the widely used  Stanford parser\footnote {http://nlp.stanford.edu:8080/parser/} as of 2014,  and the established  baseline algorithm \cite{Collins95}, which has stood the test of time.
We first manually labeled PP quads from the NYTC dataset, then prepended the noun phrase appearing before the quad, effectively creating sentences made up of 5 lexical items 
$(n0\  v\  n1\  p\  n2)$.  We then applied the Stanford parser, obtaining the results summarized in Figure~\ref{fig:parser}.  The parser performs well on some prepositions, for example,  ``of", which tends to occur with noun attaching PPs as can be seen in Figure \ref{fig:distribution}.  However, for  prepositions  with an even distribution over verb and noun attachments, such as ``on", precision is as low as ~50\%. 
The Collins baseline achieves 84\% accuracy on the benchmark Wall Street Journal PP dataset.  
 However, drawing a distinction in the precision of  different prepositions provides useful insights on its performance.  We re-implemented this baseline  and found that when we remove the trivial preposition,  ``of",  whose PPs are by default attached to the noun by this baseline, precision drops to 78\%. 
 This  analysis  suggests there is substantial room for improvement.

\subsection{Related Work}
\paragraph{Statistics-based Methods}
Prominent  prior methods learn to perform PP attachment based on corpus co-occurrence statistics,  gathered either from  manually annotated training data \cite{Collins95,BrillR94} or from  automatically acquired  training data that may be noisy \cite{Ratnaparkhi98,PantelL00}.  These models collect statistics on how often a given quadruple, $\{v,n1,p,n2\}$, occurs in the training data as a verb attachment as opposed to a noun attachment.  The issue with this approach is sparsity, that is, many quadruples occuring in the test data might not have been seen in the training data.
Smoothing techniques are often employed to overcome sparsity.  For example, \cite{Collins95} proposed a back-off model that uses  subsets of the words in the quadruple, by also keeping  frequency counts of triples, pairs and single words. 
Another approach to overcoming sparsity has been to use WordNet \cite{fellbaum98wordnet} classes, by replacing nouns with their WordNet classes \cite{Stetina97,ToutanovaMN04} to obtain less sparse corpus statistics.  Corpus-derived clusters of similar  nouns and verbs have also been used  \cite{PantelL00}.

Hindle and Rooth  proposed a lexical association approach based on how words are associated with each other \cite{HindleR93}. Lexical preference is used by  computing co-occurrence
frequencies (lexical associations) of verbs and nouns,  with prepositions. In this manner, they would discover that, for example,  the verb ``send" is highly associated with the preposition \textit{from},  indicating that in this case, the PP  is likely to be a verb attachment.

\paragraph{Structure-based Methods}
These methods are based on  high-level  observations that are then generalized into heuristics for 
 PP attachment decisions. \cite{Kimball73}  proposed a right association method, whose premise is that a word tends to attach to another word immediately to its right.
\cite{Frazier78} introduced a minimal attachment method, which posits that  words attach to an existing word using the fewest additional syntactic nodes. 
While simple, in practice these methods have been found to perform poorly \cite{WhittemoreFB90}.

\paragraph{Rule-based Methods}
\cite{BrillR94}  proposed methods that
 learn  a set of transformation rules from a corpus. 
 The rules consist of  nouns, verbs, and prepositions.
Therefore, these rules can be too  specific to have broad applicability, resulting in low recall. 
To address low recall, knowledge about nouns, as found in  WordNet,   is used to replace certain
words in rules with  their WordNet classes. 


\paragraph{Sense Disambiguation}
In addition to prior work on prepositional phrase attachment, a highly related problem is preposition sense disambiguation \cite{Hovy2011,SrikumarR13}.  Even a
syntactically correctly attached PP can  still  be semantically ambiguous with respect to questions of machine reading such as \textit{where, when,} and \textit{who}.  The same preposition
can express many semantic relations. For example, in the sentence:
 ``Poor care caused her death from pneumonia", the preposition ``from"  expresses the
relation \textit{Cause(death, pneumonia)}. But ``from" 
can denote other relations, for example in  ``copied the scene from another film"  \textit{(Source)} and  in ``recognized
him from the start (Temporal)" \cite{SrikumarR13}.
Therefore, when extracting information from prepositions, the problem of preposition sense disambiguation (semantics) has to be addressed in addition to prepositional phrase attachment disambiguation (syntax). In this paper, we consider both the syntax and semantic aspects of prepositions. 

\begin{table}[th]
\centering
\begin{tabular}{ |l| l|l|l| }
\hline
{\bf Feature Type} & {\bf \#} &  {\bf Feature}  & {\bf Example}\\
\hline

Noun-Noun Binary Relations &  & \textbf{Source: SVOs} & \\ 
   & F1. &  $svo(n2,v,n1)$  &  For q1; $(net,caught,butterfly)$\\ 
& F2. &  $\forall i : \exists sv_{i}o; $  $svo(n1,v_i,n2)$ & For q2; $(butterfly,has,spots)$ \\ 
& & & For q2; $(butterfly, can $ $see,spots)$\\ 

   Noun  Semantic Categories  &  &  {\bf Source: $\mathcal{T}$ }  & \\
                & F3. &  $\forall t_i \in \mathcal{T};$ $isA(n1,t_i)$ & For q1 $isA(butterlfy,animal)$\\ 
                   & F4. &  $\forall t_i \in \mathcal{T};$  $isA(n2,t_i)$ & For q2 $isA(net,device)$\\ 
                                                                                                          
  Verb  Role Fillers & & \textbf{Source: VerbNet} & \\ 
  & F5. &   $hasRole(n2, r_{i})$ &  For q1; $(net, instrument)$ \\

Preposition Relational  &  &  \textbf{Source: $\mathcal{M}$} & \\ 
  Definitions & F6. & $def(prep,v_i)$ $\forall i :$ &  
\\ 
         &  & $\exists sv_{i}o; v_i \in  \mathcal{M}$ $ \wedge$  &  
       \\ 
          &  &  $svo(n1,v_i,n2)$  &  For q2;
                    $def(with,has)$ \\ 
 
 Discourse Features &  & \textbf{Source: Sentence(s), $\mathcal{T}$} & \\ 
             & F7. &  $\forall t_i \in \mathcal{T}; isA(n0,t_i)$ &  $n0 \in \{n0, v, n1, p, n2\}$ \\

Lexical  Features& &  {\bf Source: PP quads} & For q1;\\ 
& F8.  & $(v, n1, p, n2)$ & $(caught,butterfly,with,net)$ \\
& F9. & $(v, n1, p)$ & $(caught,butterfly,with)$ \\
& F10. & $(v, p, n2)$ & $(caught,with,net)$ \\ 
& F11. & $(n1, p, n2)$ &$(butterfly,with,net)$ \\ 
& F12.  & $(v,p) $ & $(caught,with)$\\ 
& F13.  & $(n1,p) $ & $(butterfly,with)$\\ 
& F14. & $(p,n2) $ & $(with,net)$ \\ 
   &F15.  &$(p) $ & $(with)$\\  
\hline
\end{tabular}
\caption{Types of  features considered in our experiments. All features have values of 1 or 0.  The PP quads used as running examples are:  $q1=\{caught, butterfly, with, net\}: V$, $q2=\{caught, butterfly, with, spots\}: N$.}
\label{tab:featurevectors}
\end{table}

 \begin{table}[th]
 \centering
 \begin{tabular}{ |l| l|}

 \hline
 {\bf Prepositional Phrase Quadruple} &  {\bf Feature} \\
 \hline
 \textless Alice caught  the butterfly  with the net \textgreater: &  F1: $(net,caught,butterfly)$\\
 & F2: n/a\\
 & F3: $isA(butterlfy,animal)$\\
 & F4: $isA(net,device)$ \\
 & F5: $isA(net,insrument)$ \\
 & F6: $def(with,using)$ \\
 & F7: $isA(Alice,person)$ \\
 & F8: $(caught,butterfly,with,net)$ \\
 & F9: $(caught,butterfly,with)$  \\
 & F10:$(caught,with,net)$ \\ 
 & F11:  $(butterfly,with,net)$ \\ 
 & F12: $(caught,with)$\\
 & F13: $(butterfly,with)$\\
 & F14: $(with,net)$ \\
 & F15: $(with)$\\
 \hline
\textless The dog caught the butterfly with spots \textgreater & F1: n/a\\
& F2: $(butterfly,has,spots)$ \\
& F3: $isA(butterlfy,animal)$\\
& F4: $isA(spots,pattern)$ \\
& F5: n/a \\
& F6: $def(with,has)$ \\
& F7: $isA(dog,animal)$ \\
& F8: $(caught,butterfly,with,spots)$ \\
& F9: $(caught,butterfly,with)$  \\
& F10:$(caught,with,spots)$ \\ 
& F11:  $(butterfly,with,spots)$ \\ 
& F12: $(caught,with)$\\
& F13: $(butterfly,with)$\\
& F14: $(with,spots)$ \\
& F15: $(with)$\\
  \hline

 \end{tabular}
 \caption{Features generated for two  sentences with prepositional phrase quadruples.}
 \label{tab:featurevectorexamples}
 \end{table}

\subsection{Methodology} 

Our approach consists of first generating features from background knowledge and then training a model to learn with these features.
The trained model is applied to new sentences after also annotating them with background knowledge features. ]
 The types of features considered in our experiments are  summarized  in Table~\ref{tab:featurevectors}.  Additionally, Table \ref{tab:featurevectorexamples} shows examples of the instantiated features for two prepositional phrase quadruples.  The choice of features was motivated by our empirically driven characterization of the problem shown in Table \ref{tbl:fillerproperty}.
\begin{table}[h]
\centering
\begin{tabular}{p{7cm}}
\textit{(Verb attach) $\longrightarrow$ v $\langle$has-slot-filler$\rangle$ n2} \\
\hline
\textit{(Noun attach a.) $\longrightarrow$ n1 $\langle$described-by$\rangle$ n2}\\
\textit{(Noun attach b.) $\longrightarrow$ n2 $\langle$described-by$\rangle$ n1}\\
\end{tabular}
\caption{Explanation of verb vs. noun attachments through our empirical characterization. }
\label{tbl:fillerproperty}
\end{table}

We sampled $50$ PP quads  from the WSJ dataset. The PP quads are  labeled with noun or verb attachment. We  found that every noun or verb attachment could be explained using our threeway  characterization in  Table \ref{tbl:fillerproperty}.  
In particular, we found that in verb-attaching PPs, the second noun $n2$  is usually a role filler for the verb. Going back to our verb attaching PP in "Alice caught the butterfly with the net", we can see that the net fills the role of an instrument for the verb $catch$. On the other hand,  for noun-attaching PPs,  one noun describes or elaborates on the other. In particular, we found  two kinds of noun attachments. For the first kind of noun attachment,  the second noun $n2$ describes  the first noun $n1$, for example $n2$ might be  an attribute or property of $n1$. For example, in ``Alice caught the butterfly with the net"   the spots($n2$) are an attribute of the butterfly ($n1$).  And for the second kind of noun attachment, the first noun $n1$ describes the second noun $n2$. For example  in the PP quad $\{${\em expect, decline, in,  rates}$\}$,  where the PP ``in rates'', attaches to the $noun$. The decline:$n1$ that is expected:$v$ is in the rates:$n2$.
 We make this labeling available with the rest of the datasets.

We next describe in more detail how each type of feature is derived from the  background knowledge  in Table~\ref{tab:knowledge}.
We generate boolean-valued features for all the feature types we describe in this section.

\subsubsection{Noun-Noun Binary Relations}
The noun-noun  binary relation features, F1-2 in Table \ref{tab:featurevectors}, are boolean features   $svo(n1,v_i,n2)$ (where $v_i$ is any verb) and $svo(n2,v,n1)$ (where $v$ is the verb in the PP quad, and the roles of $n2$ and $n1$ are reversed). These features describe diverse semantic relations between pairs of nouns (e.g.,   \textit{butterfly-has-spots},   \textit{clapton-played-guitar}).  To obtain this type of knowledge, we dependency parsed all sentences in the 500 million English web pages of the ClueWeb09 corpus, then extracted subject-verb-object (SVO) triples from these parses, along with the frequency of each SVO triple in the corpus.  The value of any given feature $svo(n1,v_i,n2)$ is defined to be 1 if that SVO triple was found at least $3$ times in these SVO triples, and 0 otherwise.

To see why these relations are relevant, let us suppose that  we have  the knowledge that \textit{butterfly-has-spots}, $svo(n1,v_i,n2)$. From this, we can  infer that the PP in $\{caught,butterfly, with, spots\}$ is likely to attach to the noun.   Similarly,  suppose we  know that \textit{net-caught-butterfly}, $svo(n2,v,n1)$. The fact   that a net can be used to catch a butterfly can be used to predict that  the  PP in\\ $\{caught,butterfly, with, net\}$ is likely to attach to the verb.   

\subsubsection{Noun Semantic Categories} 
Noun semantic type features, F3-4, are boolean features   $isA(n1,t_i)$ and $isA(n2,t_i)$ where $t_i$ is a noun category in a noun categorization  scheme   $\mathcal{T}$ such as WordNet classes. 
Knowledge about semantic types of nouns, for example  that a butterfly is an animal, enables extrapolating predictions to other PP quads that contain nouns of the same type. 
We ran experiments with several noun categorizations including WordNet classes,  knowledge base ontological types,  and an unsupervised noun categorization produced by clustering  nouns based on the verbs and adjectives with which they co-occur  (distributional similarity).  

\subsubsection{Verb Role Fillers} 
The verb role feature, F5, is a boolean feature $hasRole(n2, r_{i})$ where $r_i$ is a role that  $n2$ can fulfill for the verb  $v$ in the PP quad, according to background knowledge. Notice that  if  $n2$ fills a role for the verb, then the PP is a verb attachment.  Consider the quad $\{caught,butterfly, with, net\}$, if we know that  a net can play the role of an \textit{instrument} for the verb \textit{catch}, this suggests a likely verb attachment.  We obtained background knowledge of verbs and their possible roles  from the VerbNet lexical resource ~\cite{KipperKRP08}. 
From VerbNet we obtained $2,573$ labeled  sentences containing PP quads (verbs in the same VerbNet group are considered synonymous), and the  labeled semantic roles  filled by the second noun $n2$ in the PP quad.  We use these example sentences  to label similar  PP quads, where similarity of PP quads is defined by  verbs from the same VerbNet group. 


\subsubsection{Preposition  Definitions}
The preposition  definition feature, $F6$, is a boolean feature $def(prep,v_i)=1$ $if$ $\exists v_i \in  \mathcal{M}$ $ \wedge$  $svo(n1,v_i,n2)=1$, where $ \mathcal{M}$ is a definition mapping of prepositions to verb phrases.   This mapping defines prepositions, using verbs in our ClueWeb09 derived SVO corpus, in order to capture their senses using verbs; it contains definitions such as \textit{def(with, *) = contains, accompanied by, ... }.   If  ``with" is  used in the  sense of ``contains" , then the PP  is a likely noun attachment, as in $n1$ contains $n2$ in the quad $ate, cookies, with, cranberries$. However, if  ``with" is  used in the sense of  ``accompanied by", then the PP is a likely verb attachment, as in the quad $visted, Paris, with, Sue$.

To obtain the mapping, we took the labeled PP quads (WSJ, \cite{Ratnaparkhi1994}) and computed a ranked list of verbs from SVOs, that appear frequently between  pairs of nouns  for a given preposition.
 Other   sample mappings are:  \textit{def(for,*)= used for},   \textit{def(in,*)= located in}. Notice that this feature $F6$ is a selective, more targeted version of $F2$.

\subsubsection{Discourse and Lexical Features}
The  discourse feature, $F7$,  is  a boolean feature  $isA(n0,t_i)$, for each noun category $t_i$ found in a noun category ontology  $\mathcal{T}$  such as WordNet semantic types.  For example, we might realize pick up the fact that a PP quad is surrounded by many mentions of people, or food, or organizations.  By doing this we leverage the context of the PP quad, which  can contain relevant information  for  attachment decisions. We  take into account the noun preceding  a PP quad, in particular, its semantic type. This in effect makes the PP quad into a  PP 5-tuple:  $\{n0, v, n1, p, n2\}$, where the $n0$  provides additional context.

Finally, we  use  lexical features in the form of PP quads, features F8-15.  To overcome sparsity of  occurrences of PP quads, we also use counts of shorter sub-sequences, including  triples, pairs and singles. We only use sub-sequences that contain the preposition, as the preposition has been found to be highly crucial in PP attachment decisions \cite{Collins95}.
   
\subsection{Disambiguation Algorithm}
We use the described features to train a  model for  making PP attachment decisions.
Our goal is to compute $\mathbb{P}(y|x)$,   the probability that the PP $ (p,n2)$ in the tuple $\{v,n1,p,n2\}$ attaches to the  \textit{verb (v)} , $y=1$ or  to the $noun (n1)$, $y=0$, given a feature vector $x$ describing that tuple.
As input to training the model, we are given a collection of PP quads, $D$ where  $d_i \in \mathcal{D}: d_i=\{v,n1,p,n2\} $. A small subset,
$D^l \subset \mathcal{D}$   is labeled data, thus for each $d_i \in D^l$ we know the corresponding $y_i$. The  rest of the quads, $D^u$,  are unlabeled, hence their corresponding $y_i$s are unknown.
From each PP quad $d_i$, we extract a feature vector $x_i$ according to the feature generation process  discussed earlier.. 

\subsubsection{Model}
To  model $\mathbb{P}(y|x)$, there a various possibilities. One could use a generative model (e.g., Naive Bayes) or a discriminative model ( e.g., logistic regression). In our experiments we used both kinds of models, but found the discriminative model performed better. Therefore, we present details only for our discriminative model.  We use the  logistic function: 
 \begin{equation*}
 \mathbb{P}(y|x, \vec \theta)  =  \frac{e^{\vec \theta x}}{1+ e^{\vec \theta x}}
\end{equation*}
 where $\vec \theta$  is a vector of model parameters. To estimate these parameters,  we could use the labeled data as training data and  use standard   gradient descent  to minimize the logistic regression cost function. However, we also leverage the unlabeled data.

 \subsubsection{Parameter Estimation}
 To estimate model parameters based on both labeled and unlabeled data, we use an Expectation Maximization (EM) algorithm. 
 EM estimates model parameters that maximize the expected log likelihood of the full (observed and unobserved) data.  

Since we are using a discriminative model, our likelihood function is  a  conditional likelihood function:
  \begin{align}\label{eqlikelihood}
  \mathcal{L}(\theta)  &=\sum_{i=1}^{N} \mbox{ln }  \mathbb{P}(y_i|x_i)  \nonumber \\
     &= \sum_{i=1}^{N}  y_i \theta^{T}x_i - \mbox{ln } (1+ exp ( \theta^{T}x_i)) 
   \end{align}
where $i$ indexes over the $N$ training examples.

The  EM algorithm produces parameter estimates that correspond to a local maximum in the expected log likelihood of the data under the posterior distribution of the labels, given by: \\   $\arg\max\limits_{\theta}  E_{p(y|x,\theta)} [ \mbox{ln } \mathbb{P}(y|x,\theta)]$.  In the E-step, we use the current parameters $\theta^{t-1}$ to compute the posterior distribution over the $y$ labels, give by $\mathbb{P}(y|x, \theta^{t-1})$.   We then use this posterior distribution to find the expectation of the log of the complete-data conditional likelihood, this expectation is given by  $\mathcal{Q}({\bf\theta, \theta^{t-1})}$, defined as:

  \begin{align}
\mathcal{Q}(\theta,  \theta^{t-1})  &=\sum_{i=1}^{N} E_{\theta^{t-1}} [ \mbox{ln } \mathbb{P}(y|x,\theta)]  
    \end{align}

In the M-step, a new estimate $\theta^t$ is then produced, by maximizing this $Q$ function with respect to $\theta$:
\begin{equation} \label{empameterupdate}
{\bf \theta^{t}} =\arg\max\limits_{\theta}\mathcal{Q}({\bf\theta, \theta^{t-1})}
\end{equation}

EM iteratively computes  parameters $\theta^0, \theta^1, ...\theta^{t}$,  using the above update rule at each iteration $t$, halting when there is no further improvement in the value of the $Q$ function.  Our algorithm is summarized in Algorithm 1.  The M-step solution for $\theta^t$ is obtained using gradient ascent to maximize the $Q$ function.  

 \begin{algorithm}
 \caption{The EM algorithm  for PP attachment}
 \label{algorithm}
 \begin{algorithmic}
 \STATE \textbf{Input:}  $\mathcal{X}, \mathcal{D} = D^{l} \cup D^{u}$\\
 \STATE \textbf{Output:} $\theta^{T}$
 \FOR{t = 1 . . . T}
   \STATE \textbf{E-Step:}
   \STATE Compute $p(y|x_i, \theta^{t-1})$\\
   \STATE $x_i: d_i\in D^u$; $p(y|x_i, \vec \theta)  =  \frac{e^{\vec \theta x}}{1+ e^{\vec \theta x}}$  \\
   \STATE $x_i: d_i \in D^{l}$; $p(y|x_i)= 1$ if  $y=y_i,$ else $0$ \\
   \STATE \textbf{M-Step:} 
   \STATE Compute new parameters, $\theta^{t}$\\
   \STATE ${\bf \theta^{t}}$ $=\arg\max\limits_{\theta}\mathcal{Q}({\bf\theta, \theta^{t-1})}$
   \vspace{-0.4cm}
   \begin{multline*}
    \mathcal{Q}(\theta,  \theta^{t-1})  =\sum_{i=1}^{N}\sum_{y\in\{0,1\}}   p(y|x_i,\theta^{t-1}) \times\\( y\theta^{T}x_i - \mbox {ln}  (1+ exp ( \theta^{T}x_i)) ) 
   \end{multline*}
    		
     \IF{convergence($\mathcal{L}(\theta),  \mathcal{L}(\theta^{t-1}))$}
     	\STATE \textbf{break}\\
     \ENDIF
 \ENDFOR
 \RETURN $\theta^{T}$
 \end{algorithmic}
 \end{algorithm}

\subsection{Experimental Evaluation}
We evaluated our method on several datasets containing PP quads of the form $\{v,n1,p,n2\}$.   The task is to predict if the  PP ($p,n2$) attaches to the verb $v$ or to the first noun $n1$.

\subsubsection{Experimental Setup}\label{experimentalsetup}

\begin{table}[t]
\centering
\begin{tabular}{|l|l|l|}
\hline
{\bf DataSet} &  {\bf  \# Training quads} & {\bf \# Test quads} \\
\hline
\multicolumn{3}{|c|}{Labeled data} \\
    \hline
WSJ &  20,801 & 3,097 \\
NYTC & 0 & 293  \\
 WKP & 0 & 381  \\
      \hline
 \multicolumn{3}{|c|}{Unlabeled data} \\
      \hline
          WKP  & 100,000 & 4,473,072 \\
         
\hline
\end{tabular}
\caption{Training and test datasets used in our experiments. }
\label{tab:datasets}
\end{table}

\paragraph{Datasets}
Table~\ref{tab:datasets} shows the datasets used in our experiments. 
As labeled training data, we used the Wall Street Journal  (WSJ) dataset. For the unlabeled  training data, we extracted  PP quads from Wikipedia (WKP) and randomly selected $100,000$ which we found  to be a sufficient amount of unlabeled data.
The largest labeled test dataset is  WSJ
but it is also made up of a large fraction, of  ``of" PP quads, 30\% , which trivially attach to the noun, as already seen  in Figure \ref{fig:distribution}.
 The New York Times (NYTC) and Wikipedia (WKP) datasets are  smaller but contain fewer proportions  of ``of" PP quads, 15\%,  and 14\%, respectively. 
  Additionally, we applied our  model to over 4 million unlabeled 5-tuples from Wikipedia. We make this data available for download,  along with our manually labeled NYTC and WKP datasets.
For the WKP \& NYTC corpora,  each quad has a  preceding noun, $n0$, as context, resulting in PP 5-tuples of the form:  $\{n0,v,n1,p,n2\}$. The WSJ dataset was only available to us in the form of  PP quads with no other sentence information.

\begin{table}[t]
\centering
\begin{tabular}{|l|l|l|l|l|}
\hline
& PPAD& PPAD- &Coll-& Stan-\\
& & NB &ins& ford\\
\hline
WKP &  \bf{0.793} &	0.740&	0.727&	0.701\\
\hline
WKP  & \bf{0.759} &	0.698&	0.683&	0.652 \\
\textbackslash of  & & & & \\
\hline
NYTC & \bf{0.843}	& 0.792	&0.809	&0.679\\
\hline
NYTC& \bf{0.815}	& 0.754&	0.774&	0.621\\
\textbackslash of  & & & & \\
\hline
WSJ & \bf{0.843}&	0.816&	0.841& N\textbackslash A \\
\hline
WSJ & \bf{0.779}	& 0.741&	0.778&N\textbackslash A  \\
\textbackslash of  & & & & \\
\hline
\end{tabular}
 \caption{PPAD  vs. baselines.}
   \label{fig:resultmain}
\end{table}

\begin{figure}[t]
\centering
\includegraphics[width=1\columnwidth] {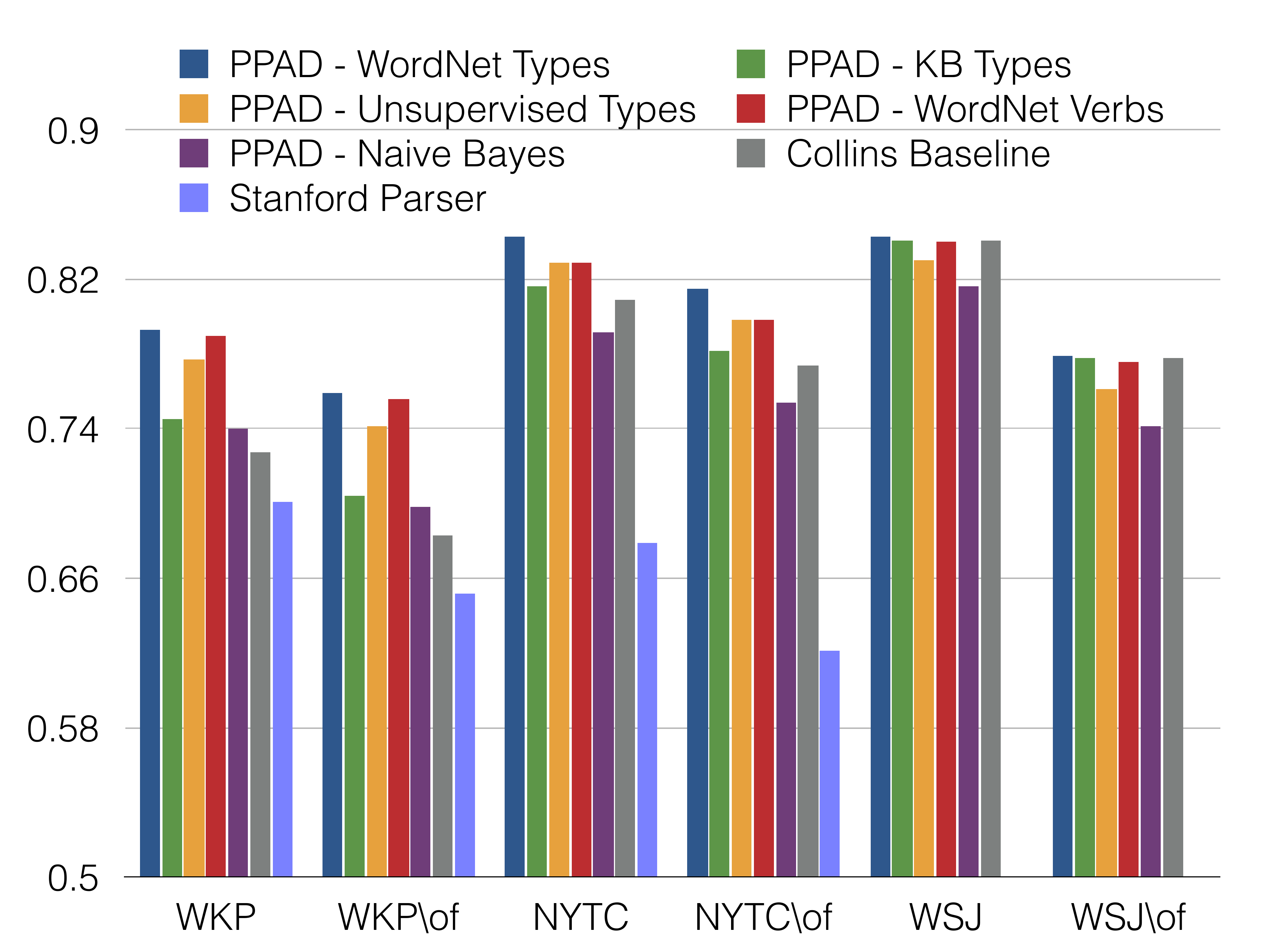}
\vspace*{-1cm}
\caption{PPAD variations vs. baselines.}
\label{fig:resultmain2}
\end{figure}

 \paragraph{Methods Under Comparison} 
\textit{1) PPAD}  (Prepositional Phrase Attachment Disambiguator) is  our proposed method. It uses diverse types of semantic knowledge, a mixture of labeled and unlabeled data for training data, a logistic regression classifier, and expectation maximization (EM) for parameter estimation \textit{2) Collins} is the established baseline among PP attachment algorithms \cite{Collins95}.  \textit{3) Stanford Parser} is a state-of-the-art dependency parser, the 2014 online version. \textit{4) PPAD Naive Bayes(NB)}  is the same as PPAD but  uses a generative model,  as opposed to the discriminative model used in  PPAD.

 \subsubsection{PPAD vs. Baselines}
Comparison results of our method to the three baselines are shown in Table \ref{fig:resultmain}. For each dataset, we also show  results when the ``of" quads are removed, shown as ``WKP\textbackslash of'', ``NYTC\textbackslash of'', and ``WSJ\textbackslash of''.
 Our method yields  improvements over the baselines. Improvements are  especially significant on the datasets for which no labeled data was available  (NYTC and WKP). On  WKP, our method is 7\% and 9\% ahead of the Collins baseline and the Stanford parser, respectively.  On  NYTC, our method is 4\% and 6\% ahead of the Collins baseline and the Stanford parser, respectively. On WSJ, which is the source of the labeled data, our method  is not significantly  better than the Collins baseline. We could not evaluate the Stanford parser on the  WSJ dataset.  The parser requires well-formed sentences which we could not generate from the WSJ dataset as it was only available to us in the form of  PP quads with no other sentence information.  For the same reason,  we could not generate discourse features,$F7$, for the  WSJ PP quads.  For the  NYTC and WKP datasets, we generated well-formed short  sentences containing only the PP quad and the noun preceding it.

 \subsubsection{Feature Analysis}
We found that features $F2$ and $F6$ did not improve performance, therefore we  excluded them from the final model, PPAD. This means that binary noun-noun relations were not useful when used permissively, feature $F2$, but when used selectively, feature $F1$, we found them to be useful. Our attempt at mapping prepositions to  verb definitions produced some noisy mappings, resulting in feature $F6$  producing mixed results.
To analyze the impact of the unlabeled data, we inspected the  features and their weights as produced by the PPAD model. From the unlabeled data, new  lexical features were discovered  that were not in the original labeled data.   Some sample  new features with high weights for verb attachments are: \textit{ (perform,song,for,*), 
(lose,*,by,*),  (buy,property,in,*)}. And for noun attachments: \textit{(*,conference,on,*), (obtain,degree,in,*), (abolish,taxes,on,*).} 

We evaluated several variations of PPAD, the results are shown in 
Figure \ref{fig:resultmain2}. For ``PPAD-WordNet Verbs",  we expanded the data by  replacing verbs in PP quads with synonymous WordNet verbs, ignoring verb senses.  This resulted in more instances of features F1, F8-10, \& F12. 

We also used  different types of noun categorizations: WordNet classes, semantic types from the NELL
knowledge base \cite{MitchellCHTBCMG15} and unsupervised types.  The KB types and the unsupervised types did not perform well, possibly due to the noise found in these categorizations.  WordNet classes showed the best results, hence they were used in  the final PPAD model for   features F3-4 \& F7.  In Section \ref{experimentalsetup}, PPAD corresponds to the best model.

\subsubsection{Application to Ternary Relations} \label{ternary}
Through the application of ternary relation extraction,  we further tested PPAD's PP disambiguation accuracy and  illustrated its usefulness for knowledge base population.
Recall that a PP  5-tuple of the form $\{n0, v, n1, p, n2\}$, whose enclosed PP   attaches to the verb $v$,  denotes a ternary relation with  arguments \textit{n0, n1, \& n2}.
Therefore, we can extract a ternary relation from every 5-tuple for which our method predicts a verb attachment.   If we have a mapping between verbs and binary relations from a knowledge base (KB), we can extend KB relations to ternary relations by augmenting the KB relations with a third argument $n2$.

\begin{table*}[ht]
\centering

\begin{tabular}{lp{1.4cm}lp{5.8cm}}
\hline
{\bf  Relation} &  {\bf  Prep. } & {\bf  Attachment accuracy}   &{\bf Example(s) } \\
\hline
acquired
& from  & \textbf{99.97}  
& BNY Mellon	\textit{acquired}	Insight 	\textit{from}	Lloyds.\\
\hline
hasSpouse & in  & \textbf{91.54} & David \textit{married}	Victoria	\textit{in}	Ireland.  \\
\hline
	worksFor  & as & \textbf{99.98} & 
Shubert	\textit{joined}	CNN	\textit{as}	reporter. \\
		\hline
playsInstrument   & with & \textbf{98.40}  & Kushner	\textit{played}	 guitar	\textit{with}	rock band Weezer. \\
\hline
\end{tabular}
  \caption{Binary relations extended to ternary relations by mapping to verb-preposition pairs in PP 5-
tuples. PPAD predicted verb attachments with accuracy \textgreater 90\% in all relations.}
   \label{tab:tenary}
  \end{table*}
      
We considered four NELL KB binary relations and their instances. For example, one instance of the $worksFor$ relation is $worksFor(Tim Cook, Apple)$. We then took the collection of 4 million 5-tuples  that we extracted from Wikipedia, and mapped verbs in 5-tuples to KB relations, based on  overlap in the instances of the KB relations, noun pairs such as $(Tim Cook, Apple)$ with the  $n0,n1$ pairs in the Wikipedia PP 5-tuple collection. We found that, for example,  instances of the noun-noun KB relation ``worksFor" match $n0,n1$ pairs in tuples where  $v= joined$ and  $p=as$ , with  $n2$ referring  to the job title (e.g., Shubert	\textit{joined}	CNN	\textit{as}	reporter ).  Other binary relations extended are: ``hasSpouse" extended by ``in" with wedding location (e.g., ``David \textit{married}	Victoria	\textit{in}	Ireland'). Or  ``acquired" extended by ``from" with the  seller of the company being acquired (e.g., BNY Mellon	\textit{acquired}	Insight 	\textit{from}	Lloyds. ).   Examples are  shown in Table \ref{tab:tenary}. 
In all these mappings, the proportion of verb attachments in the corresponding PP quads is significantly  high ( $ > 90\%$). PPAD is overwhelming  making the right attachment decisions in this setting.

Efforts in temporal and spatial relation extraction have shown that higher N-ary relation extraction is challenging.  Since prepositions specify details that transform binary relations to higher N-ary relations, our method can be used to read information that can augment  binary relations already in KBs. As future work, we would like to incorporate our method into a pipeline for reading beyond binary relations. One possible direction is to read details about the \textit{where,why, who} of events and relations, effectively moving from extracting only binary relations to reading at a more general level.

\subsubsection{Labeled Ternary Arguments}
In the above experiment, we studied the case of extending existing KB relations to ternary relations. However, we did not provide any semantic information about the role of the third arguments. In this section, we  study the case when we want to label the role of the third argument. For example, for the acquisition instance of ``BNY Mellon acquired Insight from Lloyds",  we want to predict that the label of ``Lloyds" is the ``Source",  indicating the source company of acquisition.  As another example, consider the sentence: `Bailey	bought	earrings	for	Josie",  we want to predict that the label of ``Josie" is ``Beneficiary", indicating the beneficiary of the items bought. 

To obtain labels for ternary relations we make use of VerbNet \cite{KipperKRP08}. VerbNet provides, for each verb, frames of the different use cases of the verb. Here we consider only  verb uses  that make use of prepositions. In VerbNet, these frames are described using a  label of the form: ``primary=NP V NP PP.label" where the ``label" is the role of the  argument to  the right of the prepositional phrase. One example of a VerbNet frame is:  ``primary=NP V NP PP.instrument", each such frame is accompanied by an example sentence. In this case the example is: ``Paula hit the ball  with a stick", where the ``stick" takes the role of the instrument. Notice  that a given  verb and preposition combination does not necessary invoke a given label. For example in ``Paula hit the ball with joy", ``joy" does not play the role of the instrument. Therefore,  we  introduce further constraints. We learn these constraints from the collection of 4 million 5-tuples  that we extracted from Wikipedia as explained in Section \ref{experimentalsetup}. In particular, we  replace mentions of entities with their NELL and WordNet semantic types. Using this approach, we generate templates of the form:\\
\textless np\_v\_np\_pp.LABEL \textgreater \textless verb\textgreater \textless typeofArg1\textgreater \textless preposition \textgreater \textless typeofArg2\textgreater.

We used  five labels from VerbNet: 
np\_v\_np\_pp.beneficiary, np\_v\_np\_pp.instrument,\\ np\_v\_np\_pp.asset,
 np\_v\_np\_pp.source, and  np\_v\_np\_pp.topic.  
The labels form  ternary relations as follows. Consider the sentence ``Paula hit the ball  with a stick". This sentence matches the label  np\_v\_np\_pp.instrument. The binary relation is: \textit{ hit(Paula, ball)}. Extending this to a ternary relation we get: \textit{ hit\_pp.instrument(Paula, ball, stick)}. Table \ref{tab:ternarylabelsBeneficiay} shows an example of the ternary relation label: np\_v\_np\_pp.beneficiary.
 Additional examples of learned templates for each of the five labels are as shown below in Table \ref{tab:ternarylabelsTemplates}.
 
  \begin{table}[h]
  \centering
      \begin{tabular}{l| L{10cm}}
        \hline
        
 Template:  & \textless np\_v\_np\_pp.beneficiary \textgreater \textless buy\textgreater \textless jewelry\textgreater \textless for \textgreater  \textless person\textgreater \\
\hline
 Sentence: & Sue bought earrings for Mary \\
 \hline
  Buyer: & Sue\\
  Items: & earrings\\
  Beneficiary: & Mary\\
    \hline
   Binary Relation: & buy(Sue,earrings) \\
    Ternary Relation: & buy\_pp.beneficiary(Sue,earrings,Mary ) \\
    \hline
      \end{tabular}
      \caption{Ternary Relation extracted using the label np\_v\_np\_pp.beneficiary }
        \label{tab:ternarylabelsBeneficiay}
      \end{table}

 \begin{table}[h]
 \centering
     \begin{tabular}{ll}
       \hline
       
 
\textless np\_v\_np\_pp.instrument \textgreater \textless shoot\textgreater \textless person\textgreater \textless with \textgreater \textless weapon\textgreater \\

\textless np\_v\_np\_pp.asset \textgreater \textless sell\textgreater \textless company\textgreater \textless for \textgreater \textless amount\textgreater \\

\textless np\_v\_np\_pp.source \textgreater \textless buy\textgreater \textless organization\textgreater \textless from \textgreater \textless organization\textgreater \\

\textless np\_v\_np\_pp.topic \textgreater \textless ask\textgreater \textless person\textgreater \textless for \textgreater \textless advice\textgreater \\
  
  \textless np\_v\_np\_pp.topic \textgreater \textless ask\textgreater \textless person\textgreater \textless for \textgreater \textless divorce\textgreater \\
            \hline
     \end{tabular}
     \caption{Examples of learned templates for labeled ternary relations}
       \label{tab:ternarylabelsTemplates}
     \end{table}
     
Table \ref{tab:ternarylabels}  shows sample instances of the different learned  templates for labeled ternary arguments.
We randomly sampled 100 such instances evaluated them for accuracy, we found a sampling accuracy of 88\%.

  \begin{table}[h]
 \centering
     \begin{tabular}{ll}
        \hline
        Ternary argument label &  Instance \\
       \hline
 np\_v\_np\_pp.beneficiary &	danai udomchoke	won	gold medal	for	thailand	\\
np\_v\_np\_pp.beneficiary &	alton	cooked	breakfast	for	crew	 \\
np\_v\_np\_pp.beneficiary & 	boys	cooked	cakes	for	girls	 \\
np\_v\_np\_pp.beneficiary &	bailey	buys	earrings	for	josie	 \\
np\_v\_np\_pp.beneficiary &	jim	buys	bracelet	for	kathy	 \\
np\_v\_np\_pp.beneficiary &	leonard	buys	engagement ring	for	michelle	 \\
np\_v\_np\_pp.beneficiary &	headmaster	bought	goggles	for	children	 \\
np\_v\_np\_pp.instrument &	lord edward thynne	shot	golden eagle	with	rifle	 \\
np\_v\_np\_pp.instrument &	mohawks	opened	fire	with	gunshots	 \\
np\_v\_np\_pp.instrument &	unidentified militants	opened	fire	with	grenade launcher	 \\
np\_v\_np\_pp.instrument &	jarvis	opened	fire	with	5-inch guns	 \\
np\_v\_np\_pp.instrument &	prince	stabs	vizier	with	dagger	 \\
np\_v\_np\_pp.instrument &	isaac van scoy	killed	british soldier	with	pitchfork	 \\
np\_v\_np\_pp.instrument &	ambush positions	opened	fire	with	mortars	 \\
np\_v\_np\_pp.instrument &	tamalika karmakar	killed	rebecca	with	knife	 \\
np\_v\_np\_pp.source & telugu film homam	drew	inspiration	from	martin scorsese \\
np\_v\_np\_pp.source&	john coltrane	received	call	from	davis	 \\
np\_v\_np\_pp.source&	kenneth o'keefe	received	letter	from	state department	 \\
np\_v\_np\_pp.source&	tony	receives	letter	from	mandy	 \\
np\_v\_np\_pp.source&	peter	receives	call	from	claire	 \\
np\_v\_np\_pp.source&	huppertz	drew	inspiration	from	richard wagner	 \\
np\_v\_np\_pp.source&	fiz	receives	call	from	alan hoyle	 \\
np\_v\_np\_pp.source&	elbaz	drew	inspiration	from	bruce willis	 \\
np\_v\_np\_pp.source&	smolensky	bought	company	from	wheeler	 \\
n 
np\_v\_np\_pp.topic&	wittenberg	asked	jan kazimierz	for	permission	 \\
np\_v\_np\_pp.topic&	brando	asked	john gielgud	for	advice	 \\
np\_v\_np\_pp.topic&	lutician delegates	asked	conrad	for	help	 \\
np\_v\_np\_pp.topic&	logan	asked	scott	for	help	 \\
np\_v\_np\_pp.topic&	philadelphia quakers	asked	nhl	for	permission	 \\
np\_v\_np\_pp.topic& steven	asks	frank	for	advice	 \\
np\_v\_np\_pp.topic&	rowe	asked	jackson	for	divorce	 \\
            \hline
     \end{tabular}
     \caption{Sample instances of    templates learned for labeled ternary arguments. For each instance, the label applies
     to the last argument.}
       \label{tab:ternarylabels}
     \end{table}

\subsection{Prepositional Phrase Attachment Ambiguity  Summary}
We have presented a knowledge-aware  approach to prepositional phrase (PP) attachment disambiguation, which is  a type of syntactic ambiguity. Our method incorporates  knowledge about verbs, nouns, discourse, and noun-noun binary relations.   We trained a model using both labeled data and unlabeled data, making use of expectation maximization for  parameter estimation.
Our method can be seen as an example of tapping into a positive feedback loop for machine reading enabled by recent advances in  information extraction and knowledge base construction techniques.

\clearpage
\section{Compound  Nouns Analysis} \label{nominals}
 Noun phrases contain a number of challenging compositional
phenomena, including implicit relations.
  Compound nouns such as ``pro-choice Democratic gubernatorial candidate
James Florio'', or ``White House spokesman Marlin Fitzwater'' primarily consist of  nouns and adjectives. They do not contain verbs. 
This means  that traditional pattern detection algorithms for detecting relations through lexical regularities will not work well on compound nouns.   On the other hand, beliefs such as a person’s job title, nationality, or stance on a political issue
are often expressed using compound nouns. We  propose a knowledge-aware algorithm for extracting semantic relations from compound noun analysis that learns, through distant supervision, to  map fine-grained type sequences of compound nouns to the relations they express.
Consider the following compound nouns.

\begin{table}[h]
\begin{tabular}{ll}
1.a) Giants	cornerback	Aaron Ross \\
1.b) Patriots	quarterback	 Matt Cassel \\
1.c) Colts	receiver	Bryan Fletcher \\
\\
2.a)  Japanese	astronaut	Soichi Noguchi \\
2.b) Irish	golfer	Padraig Harrington \\
2.c) French	philosopher	Jean-Paul Sartre \\
 \\
2.a) Seabiscuit	author	Laura Hillenbrand \\
2.b) Harry porter author J.K Rowling \\
2.c) Walking the Bible author Bruce Feile \\
\end{tabular}
\label{tab:nominalsexamples}
\end{table}

The concepts in the  compound noun sequences \textit{(1a. -- c.)}, \textit{(2a. -- c.)}, \textit{(3a. -- c.)}  are
of the semantic type sequences:

\begin{table}[h]
 \begin{tabular}{ll}
\textit{\textless sportsteam\textgreater \textless sportsteamposition\textgreater \textless athlete\textgreater}  \\
\textit{\textless country\textgreater \textless profession\textgreater \textless person\textgreater} \\
\textit{\textless book\textgreater  ``author" \textless person\textgreater} \\
 \end{tabular}
   \label{tab:nominalstypesequences}
 \end{table}
     
Therefore, our task is to learn semantic type sequences and their mappings to knowledge base relations. We use relations from the NELL knowledge base. Since  NELL has   binary relations that  take only two arguments, and compound nouns contain more than two noun phrases,  we additionally keep track of the position information for the two arguments of the relation.  For example, from the type sequence:   \textit{\textless country \textgreater \textless profession\textgreater \textless person\textgreater}, we generate mappings to two different  relations.

\begin{table}[h]
\begin{tabular}{llll}
\textbf{Relation}&  \textbf{arg1\_pos} &  \textbf{arg2\_pos} & \textbf{type sequence} \\
\hline
citizenofcountry & 3 & 1 & \textit{\textless country\textgreater \textless profession\textgreater \textless person\textgreater} \\
personhasjobposition & 3 & 2 & \textit{\textless country\textgreater \textless profession\textgreater \textless person\textgreater} \\
\end{tabular}
\caption{Learned mappings from compound nouns semantic type sequences to binary relations }
\label{tab:nominalsmappings}
\end{table}

\begin{figure}[t]
\centering
\includegraphics[width=1\columnwidth] {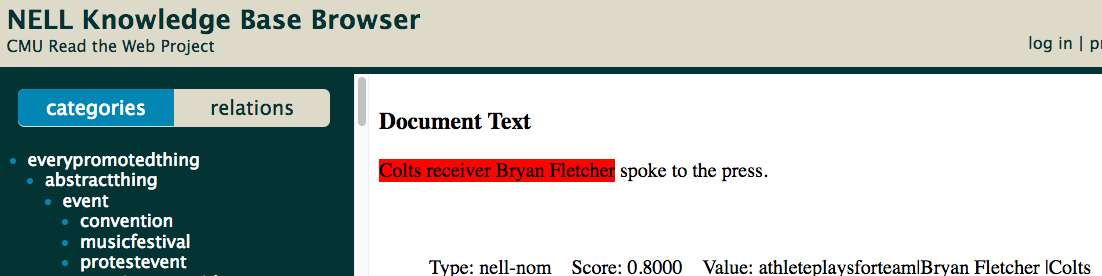}
%
\caption{Extracting the athleteplaysforteam relation from a compound noun.}
\label{fig:athletenominal}
\end{figure}

To learn  mappings from  compound nouns to binary relations as shown in Table \ref{tab:nominalsmappings}, we  use  distant supervision, that is using the NELL knowledge base as the only form of supervision.   In general, the intuition behind
 distant supervision is that a sentence that contains a pair of entities that participate in a known
 relation is likely to express that relation. In our case, the sentence is just a compound noun, for example ``Japanese	astronaut	Soichi Noguchi". Therefore, we first extract compound nouns from a large collection of documents.  For every compound noun,  we map its noun phrase to entities in the NELL. The entities are then replaced by their NELL types. This creates  type sequences of the form:  \textit{\textless country \textgreater \textless profession\textgreater \textless person\textgreater}.  Each type sequence has a support set, which is the collection of  compound nouns that satisfy the type sequence. For example, \textit{Japanese	astronaut	Soichi Noguchi} is a support compound noun   for the type sequence: \textit{\textless country\textgreater \textless profession\textgreater \textless person\textgreater}.  We retain type sequences whose support set sizes are above a threshold of  $10$ in our experiments.  For  type sequences with support set size above the threshold, we use their support sets to learn mappings from type sequences to relations using distant supervision.  That is, from each supporting compound noun we collect pairs of entities, and do look ups in NELL to determine which relations hold between the pair of entities. We additionally keep track of the position of  the entities within the compound noun. This gives us mappings from types sequences to relations such as:  \textit{\textless citizenofcountry\textgreater \textless3\textgreater  \textless1\textgreater \textless country\textgreater \textless profession\textgreater}. We only retain mappings that have a support set size (relation instances in NELL) above a threshold of $10$ in our experiments.

\subsection{Experimental Evaluation}
We extracted compound nouns from three different corpora: the New York
Times archive which includes about 1.8 Million
 articles from the years 1987 to 2007, the English edition of Wikipedia  with about
about 3.8 Million articles, and the KBP dataset \cite{conf/tac/Surdeanu13} which contains over 2 million documents with  Gigaword newswire  and Wb documents. We extracted a total of $2,270,487$ compound nouns.
From these compound nouns, we  extract 10 relations that are expressed by compound nouns and are in the NELL knowledge base. From this dataset we learned 291 mappings from  types sequences to relations.  Using these mappings we then predicted new relation instances. We report  recall and accuracy   in  Table \ref{tab:nominalsresult}. We compare our system $K-nom$ to a baseline which is not knowledge aware.  The baseline is created as follows:  we  generate  sequences that have no awareness of  background knowledge by discarding type information from the sequences. For example, the sequence \textit{\textless book\textgreater  ``author" \textless person\textgreater}  becomes \textit{\textless noun phrase\textgreater  ``author" \textless noun phrase\textgreater} where ``noun phrases" refers to any noun phrase found in text regardless of its type. In cases where the entire template is made up of semantic types only, for example, \textit{\textless country \textgreater \textless profession\textgreater \textless person\textgreater}, discarding semantic types results in a template which is too general. Therefore, for the baseline,  we discarded such permissive  templates that  would be present a significant disadvantage for the baseline.

As shown in Table \ref{tab:nominalsresult}, K-nom has high accuracy across all the relations in comparison to the baseline. K-nom also achieves high recall for some of the relations.  The reason why K-nom yields low recall   for some of the relations ia probably  because while those relations are occasionally  expressed using compound nouns, they are more commonly expressed in other forms such as using verbs.

We have incorporated K-nom into the NELL reading software. A screenshot of extracting relations from compound nouns is shown in Figure \ref{fig:athletenominal}.


\begin{table}[h]
\centering
 \begin{tabular}{|ll |l|l|}
 \hline
 Relation &  & Recall & Precision \\
   \hline
 citizenofcountry  & &  & \\
& K-nom  & 15,805 &$0.982\pm0.018$ \\
& Baseline & 22, 5453 & $0.2\pm0.08$  \\
\hline
citylocatedincountry & & & \\
&  K-nom   & 1,521 & $0.75\pm0.083$ \\
& Baseline &  544  &  $0.298\pm0.088$ \\
\hline
athleteplaysforteam & & &\\
& K-nom &  471 & $0.982\pm0.018$\\
& Baseline & 0 & 0 \\
\hline
persongraduatedfromuniversity & & & \\
& K-nom  & 49 & $0.964\pm0.036$ \\
& Baseline & 1,756 & $0.49\pm0.096$\\
\hline
personhasjobposition & & & \\
 & K-nom & 511,937 & $0.837\pm0.07$ \\
 & Baseline & 23,138 & $0.943\pm0.041$\\
 \hline
musicianplaysinstrument & & & \\
& K-nom & 3,890 & $0.885\pm0.06$ \\
& Baseline & 1,202 & $0.895\pm0.057$ \\
\hline
worksfor & & & \\
& K-nom  & 75,757 &  $0.847\pm0.068$ \\
& Baseline & 25,024 & $0.664\pm0.091$ \\
\hline
athletewinsawardtrophytournament & & & \\
& K-nom & 92 & $0.928\pm0.049$ \\
& Baseline & 6,310 &	$0.625\pm0.093$ \\
\hline
coachesteam & & & \\
& K-nom & 175 & $0.982\pm0.018$ \\
& Baseline  & 9,166 & $0.548\pm0.096$ \\
\hline
companysubsidiary& & & \\
& K-nom & 37 & $0.953\pm0.047$ \\
& Baseline  & 1,793 &  $0.423\pm0.095$ \\
   \hline
 \end{tabular}
 \caption{Comparision of our approach K-nom, to a baseline that does not make use of background knowledge. Precision is sampled
 from a total of max(100, recall). Recall is not shown in percentages as we do not have a  gold standard for recall on the corpus.}
   \label{tab:nominalsresult}
 \end{table}
    
\subsection{Related Work}
Much of the work on information extraction has been on extracting relations expressed by verb phrases
that occur between pairs of noun phrases. Extracting knowledge base relations from noun phrases alone has been much less
explored. In \cite{conf/emnlp/YahyaWGH14}, a method is  developed that learns noun phrase
structure for open information extraction. This is different from our work in that we are extract 
knowledge base relations as opposed to open information extraction. Therefore, the authors
do not ground their extracted attributes to an
external knowledge base.

The work of \cite{choi2015scalable} developed a semantic parser for extracting relations
 from noun phrases. Given an input noun phrases, it is first transformed to a logical form, where the logical form is an 
 intermediate unambiguous representation of the noun phrase.
The  logical form is chosen such that it  closely matches the linguistic structure
of the input text noun phrase. The logical form is then transformed into one that, where possible,
uses the Freebase ontology predicates \cite{Bollacker2008}. These predicates can then be read off as relations expressed about the 
entities described by the noun phrase. The authors test their work on Wikipedia category names. Since each Wikipedia category
describes a set of entities, by extracting relations from each category name, one learns relations about all the members of the category.
Consider the Wikipedia category \textit{Symphonic Poems by Jean Sibelius}.  An example of the knowledge base transformed  logical form for this category name would be: 

$\lambda x.composition.form(x;Symphonic$ $poems) \wedge composer(Jean$ $Sibelius; x)$ where one can now extract attributes for the entities, such as \textit{The Bard, Finlandia, Pohjola’s Daughter, En Saga, Spring Song, Tapiola, \ldots},  that 
fall under this category  in particular that for all $x$ in this category $composer(Jean$ $Sibelius; x)$ and  $composition.form(x; Symphonic$ $poems)$
where $composer$ and $composition.form$ are Freebase attributes. In generating the logical forms, several features are used that capture some background knowledge, in particular,   a number of features that enable soft
 type checking on the  produced logical form, and  features that test agreement of these
 types on different parts of the produced  logical form.
 
In a related but different line of work,  the NomBank project  \cite{conf/lrec/MeyersRMSZYG04,conf/acl/GerberC10} annotated the argument
structures for common nouns. For example, from the expression \textit{Greenspan’s replacement Ben Bernanke},
the arguments for the nominal ``replacement", are: ``Ben
Bernanke" is  ARG0 and ``Greenspan" is  ARG1. The resulting annotations
has been used as training data for work on semantic role labeling on nominals \cite{jiang2006semantic,conf/acl/LiuN07}.
Again, this work is different from our work in that no knowledge base relations are extracted.

There has  also  work on the broader topic of semantic structure of noun phrases. In \cite{conf/acl/SawaiSM15}, a method is proposed that parsers noun phrases into the Abstract Meaning Representation in order to detect the argument structures, and noun-noun relations in compound nouns.

\subsection{Compound  Noun Analysis Summary}
We have presented a knowledge-aware method for relation extraction from 
compound nouns.   Our method uses semantic types of concepts in compound noun sequences
to predict relations expressed by novel compound noun sequences containing concepts that we have not seen before. This  method can be
seen as another example of tapping into a positive feedback loop for machine reading made possible by projects that construct large-scale knowledge bases. Compound nouns are non-trivial to interpret in many different ways besides the noun-noun relations problem we addressed here.  For example, one problem that could benefit from background knowledge is that of   analyzing the internal structure of noun phrases through bracketing \cite{conf/acl/VadasC07,conf/acl/VadasC08}.  For example, in the noun phrase \textit{(lung cancer) deaths},  the task would be to determine that  \textit{lung cancer} modifies the head \textit{deaths}. Additionally, as future work we can increase our predicate vocabulary to learn more common sense type of relations from compound nouns, for example, in the noun phrase \textit{cooking pot}, we can extract the relation\text{purpose}, to mean the pot is used for cooking.

\section{Related Work} \label{relatedworkall}
While not many, there have been other approaches that make use of knowledge bases for machine reading. \cite{conf/acl/KrishnamurthyM14} introduced a method for  training a joint syntactic
and semantic parser. Their parser makes use of a knowledge base to produce  logical forms containing knowledge base predicates. However, their use of the knowledge base is limited to unary predicates for determining  semantic types of concepts. In contrast,  in this paper we make extensive use of  a   knowledge base augmented by linguistic resources and corpus statistics. This results in a huge collection of world knowledge  that our knowledge-aware methods have access to at inference time.

Understanding a piece of text requires both background knowledge and  context. Our focus in this paper is on background knowledge.   Recurrent neural networks (RNNs) including Long short-term memory  (LSTMs ) applied to language understanding  focus on managing  memory to enable the model to store and retrieve  context. For example RNNs have been applied to the task of answering queries about self-contained synthetic short stories,  and to the task of language modeling where the task is to predict the next word(s) in a text sequence given the previous
words \cite{MikolovKBCK10,SundermeyerSN12,WestonCB14,sukhbaatar2015end}.

These tasks are  treated as instances of sequence processing and words are stored in memory as they are read.
To retrieve relevant memories, 
smooth lookups are performed, whereby each memory is scored for its relevance, this may not scale well to our case where an entire knowledge base is considered. A notable exception in this line of work  is the approach of \cite{WestonCB14} which introduced  memory networks  combining RNN inference with a long-term memory.
One of their  experiments was performed on a  question answering task that requires   background knowledge  in the form  of statements stored as (subject, relation, object) triples. However,  this setting  is not a machine reading task. It is  an information retrieval task since there is no reading required to answer the questions. Instead  look ups are performed to find the triples most relevant to  the question.

\section{Discussion} \label{discussion}
In this paper, we have presented  results  that illustrate that  background knowledge is useful for machine reading.
Going forward, there  are still some open questions:  (i) have we captured all types of knowledge?
(ii) is our coverage of the current knowledge comprehensive? 
(iii) what other natural language understanding tasks could benefit from background knowledge.
(iv) how does context interact with background knowledge.

\subsection{Knowledge Breadth}  Have we captured all types of knowledge required to do inference for language understanding? Currently, there are some types of  knowledge not covered by our knowledge sources. For example, knowledge about actions is lacking in  both the subject-verb-object corpus statistics  and ontological knowledge bases. 
For example, commonsense knowledge about the actions a person can perform as opposed to the actions an  animal can perform. A person can cook, sing, and write a book while an animal cannot.  
Additionally, our sources  lack commonsense  knowledge pertaining to sound, for example which sounds are typical for which scenes.  Our sources cannot tell us if loud music is more likely  to be heard in a bar scene than in a hospital scene.  Our knowledge  sources  also lack spatial information. For example, our knowledge sources cannot tell us
that a street can be found in a city but not inside a car or a building. 
When these  these voids are filled in the knowledge sources, our methods can yield even more performance gains, and make them more applicable to more language understanding tasks.

\subsection{Knowledge Density} Knowledge found in knowledge bases is tied to a  formal  ontological representation, and is   therefore highly suited  to the kind of reasoning performed in  machine reading.   However, the mechanisms for building knowledge bases still have coverage limitations. For example, the NELL knowledge graph contains 1.34 facts per entity \cite{conf/emnlp/HegdeT15}.  This knowledge sparsity curtails the performance gains we can obtain from  knowledge-aware reading methods. To mitigate the this problem, in this paper we augmented  the ontological knowledge with corpus statistics consisting of subject-verb-object triples. While the corpus statistics are broad coverage , they are noisy, and are riddled with ambiguity that can  negatively impact performance. For that reason, we believe our approach will benefit from improvements in coverage of the much cleaner  ontological knowledge found in knowledge bases..

\subsection{Other Natural Language Understanding Tasks} In this paper we focused on the tasks of  prepositional phrase attachments and compound nouns. We believe a variety of tasks in natural language understanding  can benefit from background knowledge. In  noun phrase segmentation, coordinator  terms such as ``and, or"  introduce ambiguity. For example, if we encounter the sentence: `` my daughter likes cartoons so every Friday we watched Tom and Jerry ". It is not clear if ``Tom and Jerry '' denotes a single  name or two. However, since the context refers to cartoons,  we can use the  knowledge that ``Tom and Jerry" can  refer to  an animated film, to make the correct segmentation.   Co-reference resolution is still a difficult problem in natural language understanding.  This is because often there are many candidates of mentions that can co-refer. However, with relevant background knowledge,  some of those candidates can be ruled out, thereby improving accuracy of co-reference resolution.  Consider the sentence ``The bee landed on the flower because it wanted pollen.''  
If we know that  bees feed on pollen, we can correctly determine that ``it''   refers to the bee and not the flower.
In negation detection, consider the sentence:  ``Things would be different if Microsoft was headquartered in Texas.''  From this sentence alone, a machine reading program might incorrectly extract a relationship that  Microsoft is headquartered in Texas.
 But from the prior knowledge that Microsoft was never headquartered in Texas, we might be able to better detect the negation, in addition to the syntactic cues such as ``if''.  One direction for future work is to develop  knowledge-aware machine reading methods for additional tasks.

\subsection{Context vs. Background Knowledge} 
Understanding a piece of writing requires not only drawing upon background knowledge, but also upon discourse context.
Instead of reading each sentence of a document as a self-contained unit,  a machine reading program  needs to keep track of what has been stated in preceding sentences.  This is useful for dealing with  basic language concepts such as entity co-reference, but also for keeping track of  concepts already mentioned. Consider the sentence: ``John saw the girl with the binoculars". In the absence of context, the likely interpretation is that John used the binoculars to see the girl. However, if context suggests that there is a girl in possession of binoculars, the interpretation of the sentence changes. In the current work, we completely ignore context. Therefore, one direction for future work is explore how background knowledge interacts with context.

\subsection{Outlook} We are at a time where high impact technologies call for effective language under
standing algorithms: robotics, mobile phone voice assistants, and entertainment systems software. 
With knowledge-aware machine reading, we have the ambitious goal of exploiting advances in knowledge engineering to push natural language understanding systems
toward human level performance.

Lastly, exploiting the success in  building large machine learning models consisting of millions of parameters will likely further improve results produced by our approach. This is due to the ability of such models to establish non-trivial connections between different pieces of evidence.

\acks{
This research was supported by
DARPA under contract number FA8750-13-2-0005. 
Any opinions, findings,  conclusions and recommendations expressed in this paper are the authors' and do not necessarily reflect those of the sponsor.
}


\vskip 0.2in
\nocite{*}
\bibliographystyle{theapa}
\bibliography{ppad}

\end{document}